\documentclass[10pt,twocolumn,letterpaper]{article}
\pdfoutput=1
\usepackage{cvpr}
\usepackage{times}
\usepackage{epsfig}
\usepackage{graphicx}
\usepackage{amsmath}
\usepackage{amssymb}
\usepackage{mathtools}
\usepackage{booktabs}
\usepackage{verbatim}
\usepackage{multirow}
\usepackage{enumitem}
\usepackage{flushend}
\usepackage[hyphens]{url}
\usepackage{color}
\usepackage{scalerel}

\newcommand{\newdataset}{BEST}

\usepackage[pagebackref=true,breaklinks=true,letterpaper=true,colorlinks,bookmarks=false]{hyperref}

\cvprfinalcopy 

\def\cvprPaperID{3398} 

\ifcvprfinal\fi

\begin{document}

\title{The Pros and Cons: Rank-aware Temporal Attention\\for Skill Determination in Long Videos}
\author{Hazel Doughty \qquad Walterio Mayol-Cuevas \qquad Dima Damen \\
University of Bristol, Bristol, UK\\
{\tt\small <Firstname>.<Surname>@bristol.ac.uk}}

\newcommand{\fred}{\textcolor[rgb]{0.83137254902, 0, 0}{red}}
\newcommand{\fgreen}{\textcolor[rgb]{0.26666666666, 0.66666666666, 0}{green}}
\newcommand{\forange}{\textcolor[rgb]{0.8, 0.6, 0}{orange}}
\newcommand{\fblue}{\textcolor[rgb]{0.38039215686, 0.62745098039, 1}{blue}}
\maketitle
\thispagestyle{empty}

\begin{abstract}
\vspace{-0.6em}
We present a new model to determine relative skill from long videos, through learnable temporal attention modules.
Skill determination is formulated as a ranking problem, making it suitable for common and generic tasks. 
However, for long videos, parts of the video are irrelevant for assessing skill, and
there 
may be variability in the skill exhibited throughout a video.
We therefore propose a method which assesses the relative overall level of skill in a long video by attending to its skill-relevant parts.

Our approach trains temporal attention modules, learned 
with only video-level supervision, using a novel rank-aware loss function. 
In addition to attending to task-relevant video parts, our proposed loss jointly trains two attention modules to separately attend to video parts which are indicative of higher (pros) and lower (cons) skill.
We evaluate our approach on the EPIC-Skills dataset and additionally annotate
a larger dataset from YouTube videos for skill determination with five previously unexplored tasks.
Our method outperforms previous approaches and classic softmax attention on both datasets by over 4\% pairwise accuracy, and as much as 12\% on individual tasks. We also demonstrate our model's ability to attend to rank-aware parts of the video. 
   
\end{abstract}
\vspace{-0.6em}
\section{Introduction}
\vspace{-0.2em}

Skill determination is the problem of assessing how well a subject
performs a given task.
Automatic skill assessment from video will enable us to explore the wealth of online videos capturing daily tasks, such as crafts and cooking, 
for training humans and intelligent agents - \textit{which video should a robot imitate to prepare you scrambled eggs for breakfast?}

For long videos, previous approaches make a naive assumption; the same level of skill is exhibited throughout the video, and thus skill can be determined in any (or all) of its parts~\cite{doughty2017s, parmar2017learning, sharma2014video, zhang2015relative, zia2018video}.
Take for example the task of `tying a tie'; draping the tie around the neck or straightening the tie may be uninformative when determining a subject's skill, however the way the subject crosses one side over and pushes the tie into the loop are key.
Additionally, there may be variation in skill across the video: when comparing two videos, one subject may perform better at neatly crossing the tie but worse at pulling through the loop.

\begin{figure}[t]
    \centering
    \includegraphics[width=\linewidth]{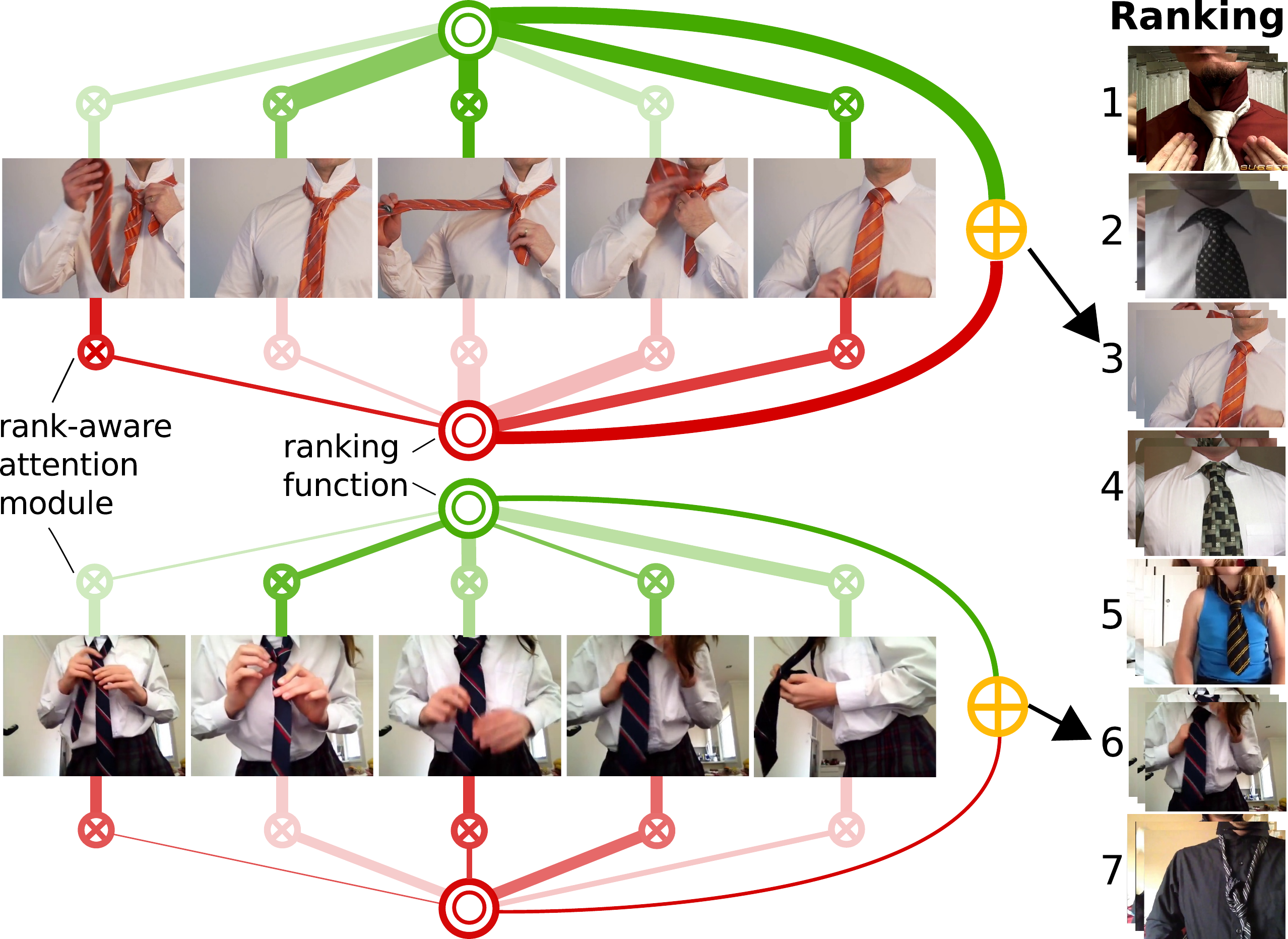}
    \caption{Rank-aware attention for skill ranking. We determine a video's rank by using high (\fgreen) and low (\fred) skill attention modules, which determine each segment's influence to the rank. Both modules are fused (\forange) for an overall skill assessment of the video. Line opacity indicates the attention value for a segment and the line thickness indicates the score.}
    \vspace{-0.6em}
    \label{fig:concept}
\end{figure}

Accordingly, we consider skill determination to be a fine-grained video understanding problem, where it is important to first localize relevant temporal regions to distinguish between instances \cite{pei2017temporal}. We target skill determination for common tasks, where ranking videos~\cite{Bertasius_2017_ICCV,doughty2017s,malpani2014pairwise} is more suitable than estimating an objective score \cite{parmar2017learning,pirsiavash2014assessing,zia2018video}.
For many tasks, objective scores would be hard to articulate or find expert bodies to certify.
Instead, crowd-sourcing can obtain a ranking on any task, which is consistent through consensus of judgment. Therefore, we devise a Siamese CNN over temporal segments, including attention modules adapted from \cite{Nguyen_2018_CVPR}, which we train to be rank-aware using a novel loss function. This is because relevance may differ depending on the skill displayed in the video - e.g. mistakes may not appear in higher-ranked videos.
When trained with our proposed loss, these modules specialize to separately attend to parts of the video informative for high skill or sub-standard performance (see Fig.~\ref{fig:concept}).

While temporal attention has previously been used to indicate relevance in long videos \cite{Nguyen_2018_CVPR, pei2017temporal}, no prior work has proposed to learn rank-aware temporal attention. 
Our \textbf{main contribution} is that we address the challenges of fine-grained video ranking by demonstrating the need for rank-aware temporal attention and propose a model to learn this effectively.
We additionally contribute a new skill determination dataset, by collecting and annotating 5 tasks from YouTube, each containing 100 videos. In total, our dataset is 26 hours of video, twice the size of existing skill determination datasets, with videos up to 10 minutes in length.
We outperform our previous effort as well as alternative attention baselines on EPIC Skills~\cite{doughty2017s} and our
newly collected dataset, BEST, and present a comprehensive evaluation of the contribution of rank-aware attention.

The rest of the paper is organized as follows. Section~\ref{sec:related} reviews the related work. We introduce our proposed method in Section~\ref{sec:method} and our new dataset in Section~\ref{sec:dataset}. 
Section~\ref{sec:results} presents quantitative and qualitative results of our method, followed by the conclusion in Section~\ref{sec:conclusion}.

\section{Related Work}
\label{sec:related}
In this section, we first review skill determination works in video, both task-specific and widely applicable methods. We then review works proposing attention modules, specifically temporal attention, for a variety of problems.

\medskip
\noindent\textbf{Skill Determination.} 
Several seminal works attempted skill determination in video \cite{gordon1995automated, ilg2003estimation, zhang2011video}. Gordon \cite{gordon1995automated} was the first to explore the viability of automated skill assessment from videos, as well as identifying appropriate tasks for analysis, with a case study on skill assessment
of gymnastic vaults from skeleton trajectories. 
Despite the importance of automatic skill assessment from video for training and guidance~\cite{Damen2014,Alayrac16unsupervised}, following works remain limited~\cite{Bertasius_2017_ICCV, doughty2017s, parmar2017learning, pirsiavash2014assessing, sharma2014video, zhang2018learning, zhang2015relative, zia2018video, zia2016automated}.
These works demonstrate good performance by focusing on features specific to the task, such as skeleton trajectory in diving~\cite{pirsiavash2014assessing} or entropy between repeated sutures in surgery~\cite{zia2018video}. 
Parallel efforts instead perform
skill determination 
from non-visual sensors such as inertial measurement units~\cite{fard2018automated, forestier2017discovering, malpani2014pairwise, wang2018deep, zia2015automated}.

Several datasets
have been introduced in prior work
\cite{doughty2017s, gao2014jhu,parmar2017learning,pirsiavash2014assessing,zhang2018learning}.
MIT Dive~\cite{pirsiavash2014assessing} and UNLV datasets~\cite{parmar2017learning} only include short video clips ($< 5$s), whilst the remaining~\cite{gao2014jhu, doughty2017s, pirsiavash2014assessing} are small scale datasets. Fis-V~\cite{zhang2018learning} contains 500 figure skating videos, however this is not publicly available. We test on our previous dataset, EPIC-Skills~\cite{doughty2017s}, as this includes the JIGSAWS~\cite{gao2014jhu} dataset re-annotated for ranking alongside 3 other tasks. We also present a new dataset for skill assessment from longer videos (avg length = 188s), consisting of 500 videos across 5 daily-living tasks.

To assess skill in long videos, different approaches have been proposed.
One is to first localize pre-selected events specific to the task~\cite{Bertasius_2017_ICCV}, such as shooting or passing the ball in a basketball game.
Alternatively, global features from the entire video have been used \cite{pirsiavash2014assessing, sharma2014video, zhang2015relative, zia2018video}, such as skeleton trajectories \cite{pirsiavash2014assessing}, features averaged across the video~\cite{parmar2017learning}, or from randomly sampled segments in our previous work~\cite{doughty2017s}.
The only work to use attention in long videos is \cite{zhang2018learning} for figure-skating. 
They use a self-attentive LSTM and a multi-scale skip LSTM to learn local (technical movements) and global (performance of players) scores respectively.
This method uses a regression framework specifically for predicting the components of figure skating scores, not appropriate for common tasks. 

We differ from all previous works in that we train a model to attend to skill-relevant parts of a video; learnable thus applicable to any task.
We use a convolutional network with temporal segments and propose a novel \textit{rank-aware} loss function.
We do not use LSTMs due to the reported issues with maintaining information over longer videos~\cite{shazeer2017outrageously, vaswani2017attention}, and inferior performance compared to non-recurrent networks in many sequence-based tasks~\cite{carreira2017quo, gehring2017convolutional, vaswani2017attention}. 

\begin{figure*}
    \centering
    \includegraphics[width=0.98\linewidth]{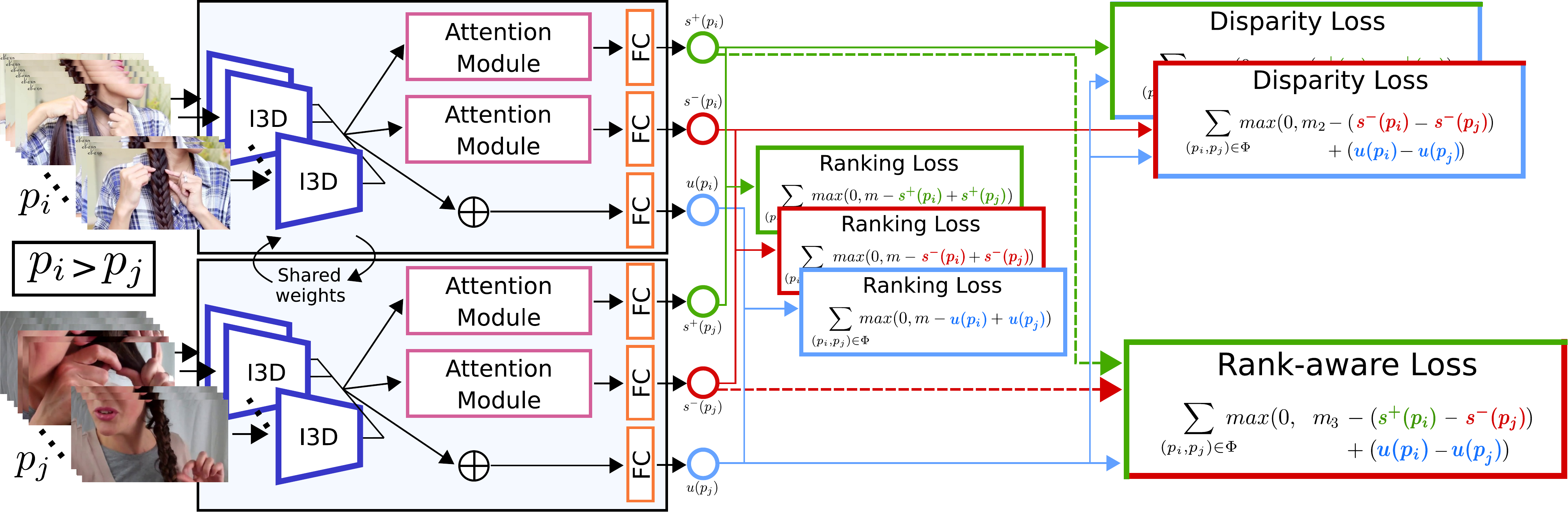}
    \caption{Rank-Aware Attention Network. Given a ranked pair of videos $(p_i, p_j)$ where $p_i$ exhibits higher skill: each video is uniformly split into segments. Extracted features (I3D) are passed into a pair of attention modules to produce video-level representations for the ranking functions (FC layers). Each ranking function produces a score $s^+$ (\fgreen) or $s^-$ (\fred). Additionally, a uniformly weighted video representation produces a third ranking score $u$ (\fblue). Three types of losses are defined: the ranking loss maximizes the margin (\fgreen-to-\fgreen, \fred-to-\fred, \fblue-to-\fblue) between the pair of ranked videos, the disparity loss ensures attention branches outperform uniform (\fgreen-to-\fblue, \fred-to-\fblue) and the final loss optimizes the attention modules to become rank-aware (\fgreen-to-\fred).}
    \label{fig:att_net}
    \vspace{-0.5em}
\end{figure*}

\medskip
\noindent\textbf{Attention Modules.}
Attention is increasingly used in fine-grained recognition, as intelligently weighting input is key to distinguishing between similar categories. 
This is a common problem in image recognition~\cite{fu2017look, singh2016end} where attention can localize discriminative attributes in the object of interest. 
For instance, Fu et al.~\cite{fu2017look} present RA-CNN to recursively zoom into the most discrimative image region with an inter-scale ranking loss. x
Singh et al.~\cite{singh2016end} adapt the spatial transformer network~\cite{jaderberg2015spatial} into a Siamese network to perform relevant attribute ranking. 
Similarly, in person re-identification from video, attention~\cite{li2018diversity, liu2017hydraplus, wu2018and} is utilized to select the frames with the best view of identifying attributes.

Attention has also been adopted in the video domain for action recognition~\cite{pei2017temporal, piergiovanni2017learning} and localization~\cite{li2018videolstm, sharma2015attention,Nguyen_2018_CVPR, Paul_2018_ECCV}, including for weakly supervised localization from video-level label
~\cite{Nguyen_2018_CVPR, Paul_2018_ECCV}. Pei et al.~\cite{pei2017temporal} combine an attention module with a gated recurrent network to classify actions in untrimmed video. 
Piergeovanni et al.~\cite{piergiovanni2017learning} present temporal attention filters to discover latent sub-events in activities. 
Nguyen et al. \cite{Nguyen_2018_CVPR} use attention filters within a CNN to identify a sparse set of video segments which minimize a video's classification loss. They use this in combination with class-specific attention from the activations to localize target actions. We build on the class-agnostic attention filters used in this work for our rank-aware attention (Sec~\ref{sec:att}). 

Using class-specific attention is a common technique in existing temporal attention works~\cite{Nguyen_2018_CVPR, Paul_2018_ECCV}. 
In this work, we propose the first model to train rank-specific (which we call rank-aware) attention, and demonstrate that it outperforms rank-agnostic attention and existing methods.

\section{Rank-Aware Attention Network}
\vspace{-0.4em}
\label{sec:method}
In this section, we re-formulate
the skill determination problem in long videos. We then detail the combination of training losses used to achieve rank-aware attention. 

\subsection{Problem Formulation}
\vspace{-0.3em}
We propose a pairwise ranking supervised learning approach for skill determination.
In this setup the training set comprises of all pairs of videos, $P$, where each pair
$(p_i, p_j) \in P$, 
has been annotated such that video
$p_i$ displays more skill than $p_j$.
Such pairwise annotations can be acquired for any task using crowd-sourcing (see Sec.~\ref{sec:dataset}). 
The aim is then to learn a ranking function $f(\cdot)$ for an individual task such that
\begin{equation}
    f(p_i) > f(p_j) \qquad \forall (p_i, p_j) \in P
\end{equation}

For long videos, \textbf{previously} we assumed these pairwise skill annotations can be propagated to any part of the video~\cite{doughty2017s}. Given $p_{it}$ is the $t^{th}$ video segment, $t \in [0,T)$, skill annotations were propagated so that, 
\begin{equation}
    f(p_{it}) > f(p_{jt}) \qquad \forall t \in [0,T); (p_i, p_j) \in P
\end{equation}

Another approach to deal with long videos~\cite{parmar2017learning, yao2016highlight}, is to use a uniform weighting of feature vectors to learn a video level ranking.
This assumes all parts of the video are equally important for skill assessment, i.e. $u(p_i) > u(p_j)$ where, 
\vspace{-0.3em}
\begin{equation} u(p_i) =  f(\frac{1}{T}\sum_t {p_i}_t)  
    \label{eq:uniform}
\end{equation}

\textbf{In this work}, we believe these assumptions do not hold. First, some parts of the video may not exhibit any difference in skill, or may even show reversed ranking - where the overall better video has segments exhibiting less skill. Second, non-uniform pooling should better represent the video's overall skill by increasing the weight for segments more pertinent to a subject's skill. 
Third, comparing corresponding video chunks $(p_{it}, p_{jt})$ assumes tasks are performed in a set order, at the same speed. We deviate from these assumptions, and
instead aim to jointly learn temporal attention $\alpha(\cdot)$, alongside ranking function $r(\cdot)$ such that
\begin{equation}
\vspace{-0.3em}
s(p_i) > s(p_j); \qquad s(p_i) = r(\sum_t \alpha({p_i}_t) {p_i}_t)
\end{equation}
While $\alpha(\cdot)$ is a standard attention module for relevance, we observe that the segments most crucial to determining skill may differ depending on the subject's skill; a low-skill subject may perform certain actions (e.g.~mistakes) not performed by a high-skill subject and vice-versa. Therefore, we propose to train two general attention modules to produce scores $s^+, s^-$, for all pairs $(p_i, p_j) \in P$, such that:
\begin{equation}
\resizebox{0.9\linewidth}{!}{
    $s^+(p_i) > s^+(p_j);$ \quad
    $s^-(p_i) > s^-(p_j);$ \quad
    $s^+(p_i) \gg s^-(p_j)$}
\end{equation}
In particular, $s^+(p_i) \gg s^-(p_j)$, encourages the
two attention modules to diverge, such that one attends to segments which display a high skill ($\alpha^+$) and the other to low skill~($\alpha^-$), along with differing ranking functions $g$, $h$:

\begin{equation}
    s^+(p_i) = g(\sum_t \alpha^+({p_i}_t) {p_i}_t)
    \label{eq:splus}
\end{equation}

\begin{equation}
    s^-(p_i) = h(\sum_t \alpha^-({p_i}_t) {p_i}_t)
    \label{eq:sminus}
\end{equation}

\subsection{Rank-Aware Attention and Overall Network}
\vspace{-0.3em}
\label{sec:net}
We show our overall architecture in Fig.~\ref{fig:att_net}. The Siamese network takes a video pair $(p_i, p_j)$ and splits each into $T$ segments of uniform length.  
The features from all segments $\{{p_i}_t\}$ are then passed to three branches.
Within each branch, we first obtain a video level representation from all segments either
weighted by our learned attention functions $\alpha^+(\cdot)$ and $\alpha^-(\cdot)$ (Sec.~\ref{sec:att}), or through uniform weighting $\frac{1}{T} \sum_t^T {p_i}_t$.
Three ranking functions are then learned (one per branch)
$g(\cdot), h(\cdot)$ and $f(\cdot)$ with a fully connected (FC) layer to produce
corresponding scores per video $s^+$ (Eq.~\ref{eq:splus}), $s^-$~(Eq.~\ref{eq:sminus}) and $u$~(Eq.~\ref{eq:uniform}). The FC layers are separate for each weighting function, but shared by both sides of the Siamese network. These scores are then evaluated by different loss types: ranking loss, disparity loss and rank-aware loss, each of which is explained below.

For each branch, a margin \textbf{ranking loss} function ensures $p_i$ is ranked higher that $p_j$,
\vspace{-0.2em}
\begin{equation}
\label{eq:rank}
    L_{rank}^+ = \sum_{(p_i, p_j) \in P} max(0, m - s^+(p_i) + s^+(p_j))
    \vspace{-0.2em}
\end{equation}
where $s^+(p_i)$ is the final score of video $p_i$ from the high-skill attention module and $m$ is a constant margin. The ranking loss is defined similarly for the low-skill and uniform weighting branches:
\begin{align}
\vspace{-0.4em}
\label{eq:rank2}
    &L_{rank}^- &= &\sum_{(p_i, p_j) \in P} max(0, m - s^-(p_i) + s^-(p_j))\\
\label{eq:rank3}
    &L_{rank}^u &= &\sum_{(p_i, p_j) \in P} max(0, m - u(p_i) + u(p_j))
    \vspace{-0.4em}
\end{align}
While the need for uniform weighting may not be obvious,
we empirically noted that ranking using the attention module frequently falls into local-minima during training. The learned attention weights for such a local-minimum perform worse than uniform weighting. 
We avoid this by introducing 
an attention \textbf{disparity loss}, which explicitly encourages an attention branch to outperform uniform: 
\begin{align}
\label{eq:net_rank}
L_{disp}^+ = \smashoperator{\sum_{(p_i, p_j) \in P}} max(0, m_2 &- (s^+(p_i) - s^+(p_j))\nonumber \\[-1em]
&+ (u(p_i) - u(p_j)))
\end{align}
Here, $m_2$ is a separate margin from $m$ specific to this loss.
For a video pair $(p_i, p_j)$, this loss encourages the difference between scores $(s^+(p_i), s^+(p_j))$ to be greater than the difference between scores $(u(p_i), u(p_j))$, thereby encouraging the attention module to produce video-level representations better at distinguishing between the skill displayed in the two videos than uniform weighting. This loss alone could instead cause the performance of $f(\cdot)$ to degrade, however by jointly optimizing with  Eq.~\ref{eq:rank3} this is avoided.
An analogous loss $L_{disp}^-$ is defined for the low-skill branch.

Using the loss functions defined
so far, the two learned attention modules $\alpha^+(\cdot), \alpha^-(\cdot)$ are indistinguishable.
They attend to skill-relevant segments to form video-level representations and $g(\cdot)$ and $h(\cdot)$ perform the ranking. 
We finally optimize these filters to achieve the desired response with our proposed \textbf{rank-aware loss}: 
\begin{align}
    \label{eq:rankAWareLoss}
    L_{rAware} = \smashoperator{\sum_{(p_i, p_j) \in P}} max(0,m_3 &- (s^+(p_i) - s^-(p_j)) \nonumber \\[-1em]
&+ (u(p_i) - u(p_j)))
\end{align}
With Eq.~\ref{eq:rankAWareLoss}, 
we ensure $s^+$ attends to higher skill parts of the better video $p_i$ while $s^-$ attends to video parts with lower skill from $p_j$. To optimize for rank-aware attention, we use a larger margin $m_3$ compared to single branches $m_2$. 
The overall training is then conducted by combining the losses:
\begin{equation}
    \label{eq:overallLoss}
     L_R = \sum_{{\scaleto{i=\{+,-,u\}}{6pt}}} L_{rank}^i + 
    \sum_{{\scaleto{i=\{+,-\}}{6pt}}} L_{disp}^i + L_{rAware}
\end{equation}

As training iterates through pairs in $P$, the same video will be considered higher skill in one pair and lower in another (e.g. $(p_i, p_j) \in P, (p_j, p_k) \in P$). The network accordingly optimizes the \textit{shared weights} so as to learn rank-aware attention modules.

When \textbf{testing} the network, a single video is evaluated and its rank is assigned through its ranking score:
\begin{equation}
    \label{eq:test}
    R(p_i) = s^+(p_i) + s^-(p_i)
\end{equation}

Note that in training we learn $s^+(\cdot)$ and $s^-(\cdot)$ such that $s^+(p_i) > s^+(p_j)$ and $s^-(p_i) > s^-(p_j)$ which implies $s^+(p_i) + s^-(p_i) > s^+(p_j) + s^-(p_j)$. Although $\alpha^-(\cdot)$ attends to low-skill segments, the overall score $s^-$ reflects the correct ranking of the videos. We do not include $u(p_i)$ as the attention alone should be sufficient (shown in Fig.~\ref{fig:branches}).

\begin{figure}[t]
    \centering
    \includegraphics[width=0.95\linewidth]{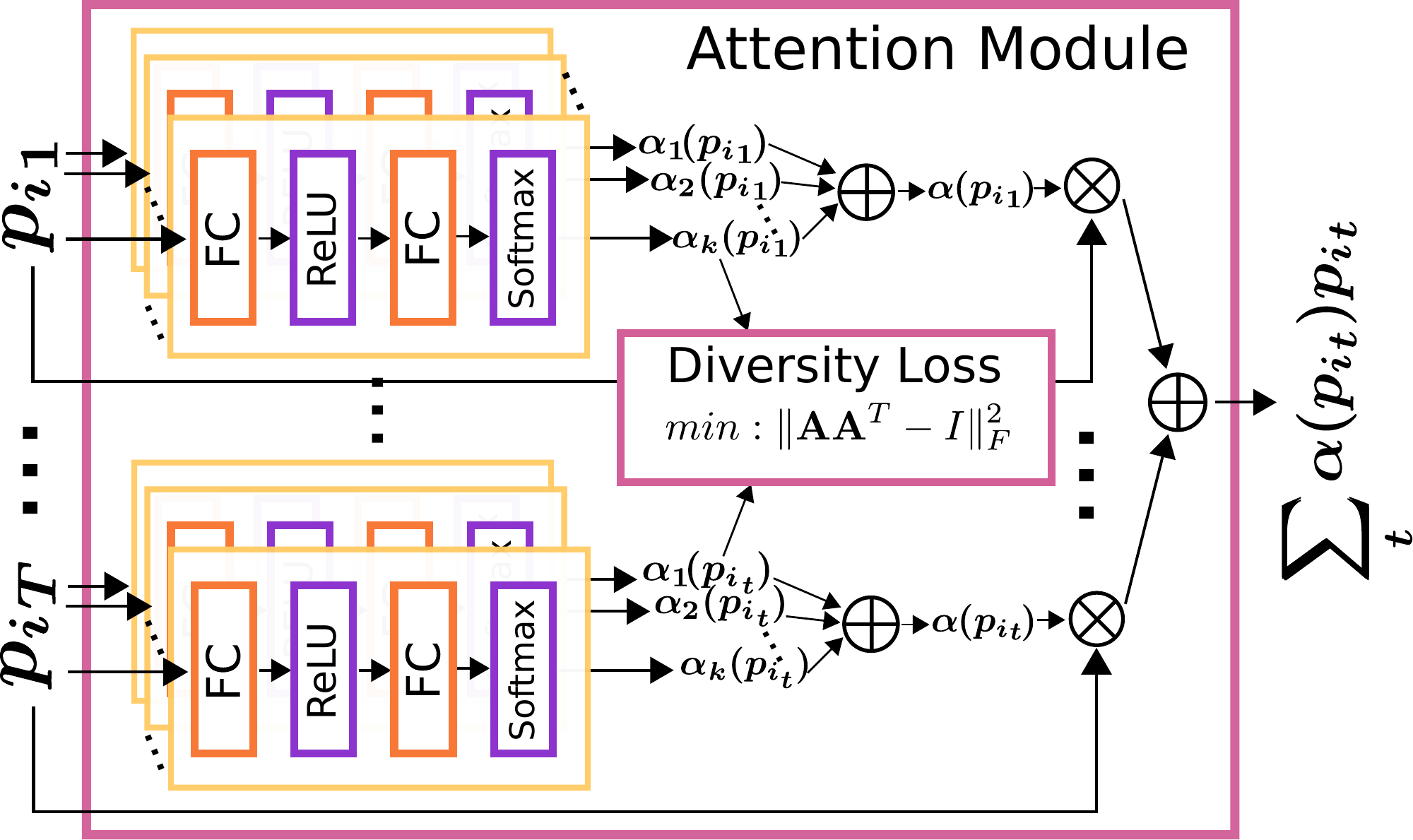}
    \caption{The attention module consists of $K$ attention filters, each outputting a scalar weight per segment, used to produce the weighted video-level feature.}
    \label{fig:att_module}
    \vspace{-0.7em}
\end{figure}

\subsection{Multi-filter Attention Module}
\vspace{-0.3em}
\label{sec:att}
Our attention modules $\alpha^+(\cdot)$ and $\alpha^-(\cdot)$ each take a set of $T$ video segments and learn a weighting of these segments informative for skill ranking. 
As the attention modules have the same structure, we will refer to the generic attention module $\alpha(\cdot)$ for simplicity. We show the architecture of the attention module in Fig.~\ref{fig:att_module}. The attention module consists of $K$ filters, each comprised of two FC layers, the first followed by a ReLU activation function, the second followed by a softmax. This is based on the attention filter used in~\cite{Nguyen_2018_CVPR} with a softmax activation instead of sigmoid. Filters are combined to achieve segment level attention:
\begin{equation}
    \alpha({p_i}_t) = \sum_{k=1}^K \alpha_k({p_i}_t)
\end{equation}
where $\alpha_k$ refers to the $k$th attention filter for the attention module $\alpha(\cdot)$, and importantly $\sum_{t=1}^T \alpha_k({p_i}_t) = 1$ for each of the $K$ filters. 
We include multiple attention filters to encourage a module to attend to multiple skill-relevant sub-tasks in the long videos; a single filter typically focuses on only one element of the task~\cite{long2018attention}. 
To regularize the $K$ filters, we use a diversity loss. We define the $K$ x $T$ attention matrix relating to video $p_i$ as:
\begin{equation}
\mathbf{A}_i = \begin{bmatrix}
    \alpha_1({p_i}_1) & \alpha_1({p_i}_2) & \dots  & \alpha_1({p_i}_t) \\
    \alpha_2({p_i}_1) & \alpha_2({p_i}_2) & \dots  & \alpha_2({p_i}_t) \\
    \vdots & \vdots & \ddots & \vdots \\
    \alpha_k({p_i}_1) & \alpha_k({p_i}_2) & \dots  & \alpha_k({p_i}_t)
\end{bmatrix}
\end{equation}
and use the following \textbf{diversity loss}:
\begin{equation}
    \label{eq:diversity}
    L_{div} = \sum_{(p_i, p_j) \in P} \|\mathbf{A}_i\mathbf{A}_i^T - \mathbf{I}\|_F^2 + \|\mathbf{A}_j\mathbf{A}_j^T - \mathbf{I}\|_F^2
\end{equation}
where $\mathbf{I}$ is the identity matrix and $\|\cdot\|_F^2$ denotes the Frobenius norm.
Similar losses have been used successfully in other applications, such as text embedding~\cite{lin2017structured} - here we use it to regularize temporal attention in video. In our network, this loss encourages each filter to learn a different aspect of the video.
Without such a loss, all filters attend to the same most discriminative part in the video, rendering more than one filter redundant. This loss also encourages filters to be sparse and pick the few most informative segments.
We assess the effect of multiple filters in Section~\ref{sec:results}.

Note that the diversity loss is within an attention module; diversity is not enforced between modules. Attentions are allowed to overlap and do so when the segment is relevant for different skill levels.
Our overall training loss is:
\begin{equation}
    L_R = \smashoperator{\sum_{{\scaleto{i=\{+,-,u\}}{6pt}}}} L_{rank}^i +
    \lambda \smashoperator{\sum_{{\scaleto{i=\{+,-\}}{6pt}}}} L_{div}^i +  
    \smashoperator{\sum_{{\scaleto{i=\{+,-\}}{6pt}}}} L_{disp}^i + L_{rAware}
    \label{eq:overall2}
\end{equation}

\section{Tasks and Datasets}
\vspace{-0.4em}
\label{sec:dataset}
We evaluate our model on our previous dataset, EPIC-Skills~\cite{doughty2017s}.
It consists of four distinct tasks: surgery (knot-tying, needle passing, and suturing) from~\cite{gao2014jhu}, dough-rolling from~\cite{de2009guide} as well as self-recorded drawing (two drawings) and chopstick-using. 
Every (sub-)task consists of up to 40 videos, with pairwise annotations indicating the ranking of videos in a pair. A limitation of this dataset is that each task is collected in a single environment with the same perspective and only minor variations in the background. We therefore collect and annotate a new skill determination dataset over twice as large, from online videos and thus with a variety of individuals, environments, and viewpoints.

\begin{table}[t]
\begin{center}
{\def\arraystretch{1.1}\tabcolsep=2.2pt
\begin{tabular}{@{}llllll@{}}
\toprule
 & Task & \#Videos & \#Pairs & \%Pairs & Av. Length (s)\\
\midrule
\multirow{4}{*}{\rotatebox[origin=c]{90}{EPIC-Skills}} & Chopstick Using & 40 & 536 & 69\% & \enskip 46 $\pm$ 17\\
& Dough Rolling & 33 & 181 & 34\% & 102 $\pm$ 29\\
& Drawing & 40 & 247 & 65\% & 101 $\pm$ 47\\
& Surgery & 103 & 1659 & 95\% & \enskip 92 $\pm$ 41\\
\midrule
\multirow{5}{*}{\rotatebox[origin=c]{90}{\newdataset}} & Scramble Eggs & 100 & 2112 & 43\% & 170 $\pm$ 113\\
& Tie Tie & 100 & 3843 & 77\% & \enskip 81 $\pm$ 47\\
& Apply Eyeliner & 100 & 3743 & 76\% & 122 $\pm$ 105\\
& Braid Hair & 100 & 3847 & 78\% & 179 $\pm$ 91\\
& Origami & 100 & 3237 & 65\% & 386 $\pm$ 193\\
\bottomrule
\end{tabular}
}
\end{center}
\vspace{-0.2em}
\caption{Comparing EPIC-Skills with BEST: \#videos, \#of pairs and average and standard deviation of video length.}
\label{tab:dataset}
\vspace{-0.6em}
\end{table}

\subsection{\newdataset~Dataset}
\vspace{-0.3em}
We collect and annotate the Bristol Everyday Skill Tasks (BEST) 2019 dataset consisting of five skill tasks with 100 videos per task, publicly available\footnote{\url{https://github.com/hazeld/rank-aware-attention-network}}. This dataset gives us an opportunity to test on a larger variety of skill tasks with more and longer videos per task from varied environments. 

\medskip
\noindent \textbf{Video Collection}.
We selected five tasks which can be completed using various methods and may be challenging for novices: scrambling eggs, braiding hair, tying a tie, making an origami crane and applying eyeliner. 
The tasks selected are deliberately varied in their content and also differ from the tasks in EPIC-Skills as this allows a more thorough testing of the proposed model.

To obtain 100 videos per task, we first retrieve the top-400 videos from YouTube using the task name as a query.
We then ask AMT workers to answer questions about each video to determine its suitability for our dataset. These ensure the selected videos contain the relevant task, are good quality videos, contain a clear view of the task and the complete performance of the task with minimal edits. We also ask AMT workers for their initial opinion of the skill of the person performing the task: `Beginner', `Intermediate' or `Expert'. This initial labelling ensures we select sufficient beginner videos before pairwise annotations.

As only a portion of the YouTube video may contain the desired task, we annotate the start and end of the relevant activity via AMT, using the same approach for annotations from~\cite{Damen2018EPICKITCHENS}. We use the agreement of 4 workers. 

\medskip
\noindent\textbf{Pairwise Annotation.}
As in~\cite{doughty2017s}, we ask AMT workers to watch videos in a pair simultaneously and select the video which displays more skill. The pair is taken as ground-truth only if all four workers agree on a pair's ordering. 
It is unnecessary to annotate all possible pairs. Instead, we annotate 40\% of the possible pairings, where each video appears in an equal number of pairs. We remove the need for exhaustive annotation by utilizing the transitive nature of skill ranking to obtain pairs outside of the original 40\%.
We then perform a second round of annotations for pairs of a similar rank, to ensure our dataset contains challenging pairs.

The number and percentage of pairs per task is shown in Table~\ref{tab:dataset}, along with the average video length per task. Our dataset is considerably larger than our previous effort EPIC-Skills in terms of both videos and annotated pairs. 

\section{Experiments}
\vspace{-0.4em}
\label{sec:results}
We first describe the implementation details of our network. We then present results on the two datasets alongside baselines and analyze the contribution of the various components in our method with an ablation study. 

\subsection{Implementation Details}
\vspace{-0.3em}
We uniformly sample 400 stacks of 16 frames, at 10fps, for each video. 
Images are re-scaled to have a height of 256 pixels then centre cropped to 224$\times$224. 
We extract features using 
I3D, pre-trained on Kinetics~\cite{carreira2017quo}. 
To prevent overfitting we augment the features by adding noise $\mathcal{N}(0, 0.01^2)$ 
per dimension as in~\cite{Nguyen_2018_CVPR}.
All models are trained using the Adam optimizer with a batch size of 128 and learning rate of $10^{-4}$ for 2000 epochs. 
For stable training, we iteratively optimize the network's parameters. 
We first fix the attention module parameters and optimize the ranking FC layer weights using $L_{rank}$ losses (Eq~\ref{eq:rank},~\ref{eq:rank2},~\ref{eq:rank3}).
We then fix the ranking FC layer weights and optimize the attention module weights, using the remaining losses ($L_{div}$, $L_{disp}$ and $L_{rAware}$).
In all experiments, we set the weight of $\lambda$ (Eq.~\ref{eq:overall2}) to 0.1, $m_1=1$ (Eq.~\ref{eq:rank}), $m_2=0.1$ (Eq.~\ref{eq:net_rank}) and $m_3=0.3$ (Eq.~\ref{eq:rankAWareLoss}).

\begin{figure*}
\centering
\includegraphics[width=\linewidth]{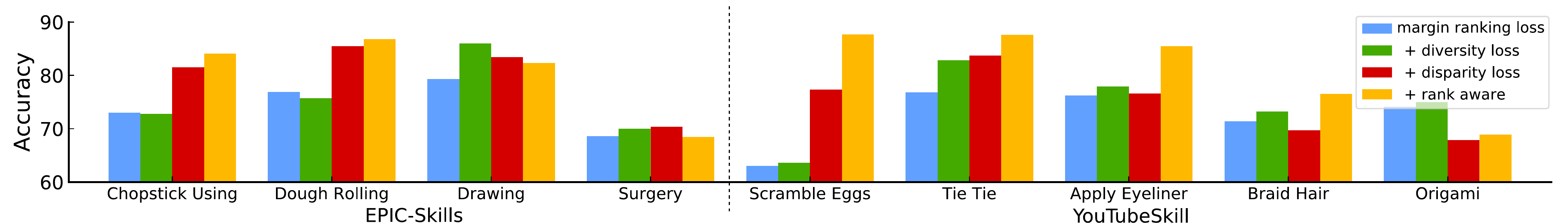}
\caption{Ablation study of loss functions on all tasks. In general each additional loss term gives an improvement, the most significant improvement being the rank-aware loss which gives an average 5\% improvement for \newdataset.}
\label{fig:ablation}
\vspace{-1em}
\end{figure*}
\begin{figure*}
\centering
\includegraphics[width=\linewidth]{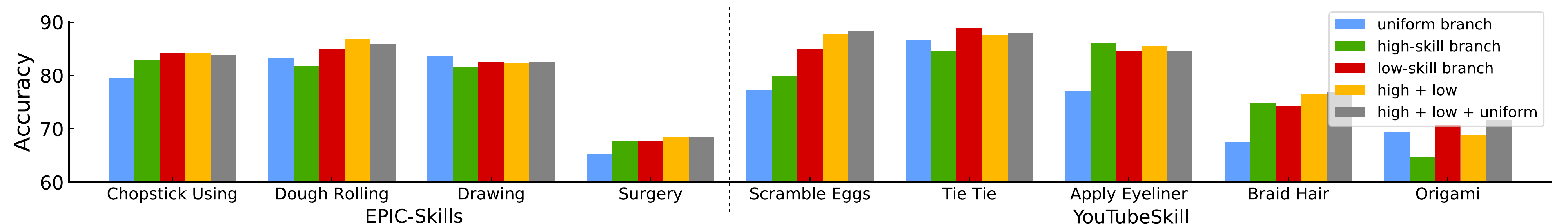}

\caption{Contribution of different branches in the network. The addition of $L_{disp}^+$ and $L_{disp}^-$ cause both the high and low skill branches to perform better than uniform in most tasks. These branches offer complementary information causing an improvement in our final result.}
\label{fig:branches}
\vspace{-0.5em}
\end{figure*}

\subsection{Quantitative Results}
\vspace{-0.3em}
\medskip
\noindent\textbf{Evaluation Metric}
We evaluate tasks individually and report pairwise accuracy (\% of correctly ordered pairs) and mean task accuracy for each dataset. For EPIC-Skills we use the four-fold cross validation training and test splits provided with the dataset~\cite{doughty2017s}. For \newdataset~we use a single 75\%:25\% split per task (provided with release), as the number of pairs is larger. 
Our test set consists exclusively of pairs where neither video is present in the training set.

\begin{table}[t]
\begin{center}
{\def\arraystretch{1.2}\tabcolsep=9pt
\begin{tabular}{@{}lll@{}}
\toprule
Method & EPIC Skills & \newdataset \\
\midrule
Who's Better~\cite{doughty2017s} & 76.0 & 75.8\\
Last Segment & 76.8 & 61.0 \\
Uniform Weighting & 78.8 & 73.6 \\
Softmax Attention & 74.5 & 72.3 \\
STPN~\cite{Nguyen_2018_CVPR} &  74.3 & 70.0\\

\midrule
Ours (Rank Aware Attention) & \textbf{80.3} & \textbf{81.2}\\
\bottomrule
\end{tabular}
}
\end{center}
\vspace{-0.5em}
\caption{Results of our method in comparison to baseline. Our final method outperforms every baseline on both datasets.} 
\label{tab:base}
\vspace{-0.5em}
\end{table}

\medskip
\noindent\textbf{Baselines and Attention.} In Table~\ref{tab:base} we show the results of our method in comparison with different baselines. 

We outperform our previous work~\cite{doughty2017s} by 4.3\% and 5.4\% on EPIC-Skills and BEST respectively.
We also use four baselines for various temporal attention approaches. The first temporal attention baseline selects only the \textbf{last segment} of the video as skill-relevant. 
It could be argued that this segment, displaying the final outcome of the task, is sufficiently informative to attend to across tasks, however this performs particularly poorly on \newdataset. 
We also use \textbf{uniform weighting} and \textbf{softmax attention} as temporal attention baselines. For softmax attention we use our method with a single attention branch only optimized by $L_{rank}$. 
Importantly, our proposed method shows an improvement over both uniform weighting and standard softmax attention, particularly for \newdataset~with longer videos. 
Interestingly, we see the inclusion of softmax attention decreases the accuracy for both datasets from a naive uniform weighting of segments (-4.3\% and -0.7\%). Although softmax attention achieves higher accuracy than uniform for several tasks, we found softmax attention to be highly inconsistent. 
To compare to existing temporal attention methods, we adapt the class agnostic attention from Sparse Temporal Pooling Network (STPN)~\cite{Nguyen_2018_CVPR} into a pairwise ranking framework. 
While this method works well for action localization, in a ranking framework it performs worse than both our method and uniform sampling.

In general the baselines struggle on \newdataset~as they are affected by the lengthy videos and increase in irrelevant parts, while
last segment is affected by variations in environment and viewpoint. By focusing on key segments indicative of skill, our method is able to combat these difficulties and gain a larger increase on this dataset.

\medskip
\noindent\textbf{Ablation Study.}
In Fig.~\ref{fig:ablation} we perform a per-task ablation study, testing the individual contributions of the components of our loss function (Eq.~\ref{eq:overallLoss}). 
The inclusion of the diversity loss increases the result by~2\% for both datasets. It is particularly useful for Drawing (+7.3\%) and Tie Tie~(+6\%), as videos in these tasks consistently have many skill-relevant segments.

\begin{figure}
\includegraphics[width=0.82\linewidth]{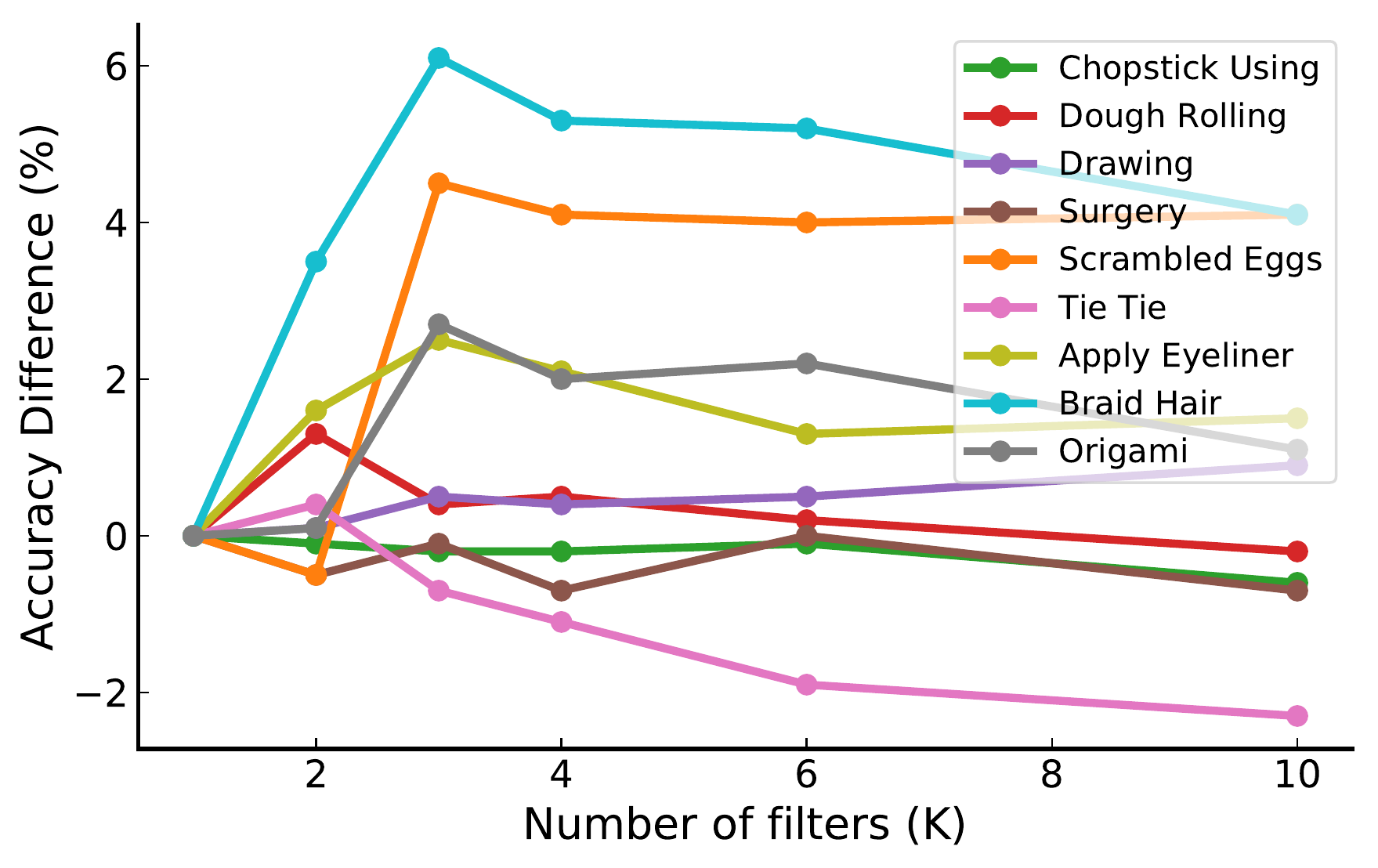}

\caption{We test the number of filters ($K$) for all tasks. The number of filters causes a clear increase in many tasks, with the majority of tasks peaking at $K=3$}
\vspace{-0.9em}
\label{fig:filters}
\end{figure}

From Fig.~\ref{fig:ablation} we see training the attention module alongside the uniform weighting with the disparity loss improves the results further. 
$L_{disp}$ encourages the network to learn attention better at discriminating between videos than the uniform weighting and decreases the sensitivity to initialization. 
In tasks like Chopstick Using and Scramble Eggs, where attention optimized with only the ranking loss performs similarly to uniform, this can help significantly. 

Our final rank-aware loss
further improves the results, particularly for \newdataset~(average improvement of 5\%). 
This is especially true for Scramble Eggs and Apply Eyeliner (+10.4\% and +8.8\% respectively). These tasks contain more instances of subtasks specific to subjects with higher or lower skill, as can be seen in Section~\ref{sec:qual}.

We note three exceptions to this trend: Drawing, Surgery and Origami. Surgery maintains a similar score throughout the ablation test and has the lowest final score of all tasks. We believe this is due to the I3D features not being able to capture the difference between the fine-grained detail of surgical motions of different abilities. Drawing and Origami both drop with the addition of $L_{disp}$. In Drawing the attention branch struggles to be better at separating videos than the uniform branch, indicating most segments are relevant for determining skill.
In Origami, the uniform weighting has poor performance due to the visual subtlety of placing neat folds in the paper. 
Therefore, optimizing the attention branch to be better than uniform does not improve training.

\begin{figure}
\centering
\includegraphics[width=0.95\linewidth]{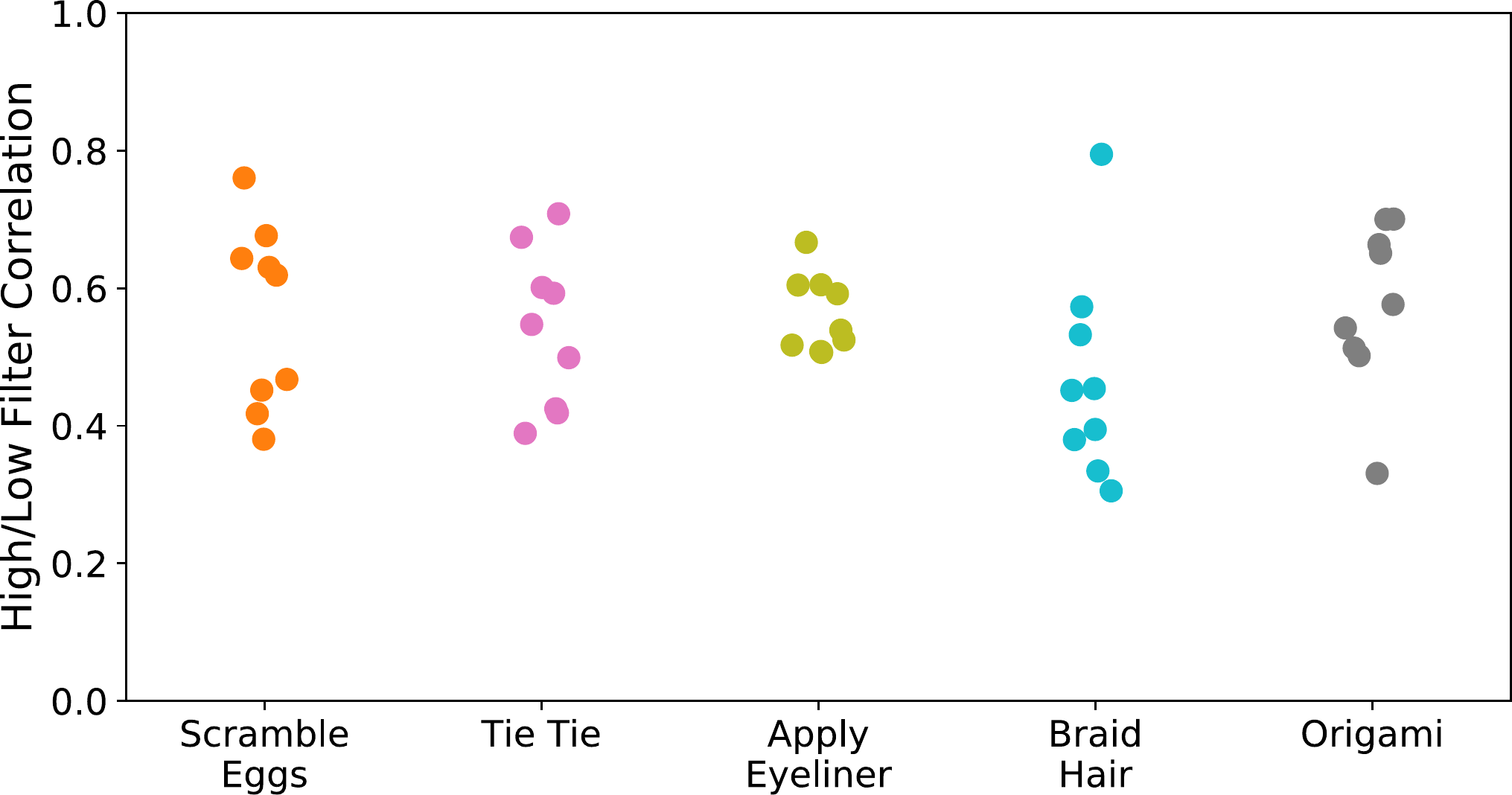}

\caption{We test correlation of high and low skill filters for all tasks, to check they attend to different video segments.}
\vspace{-0.6em}
\label{fig:correlation}
\end{figure}

\begin{figure*}
    \centering
    \includegraphics[width=\linewidth]{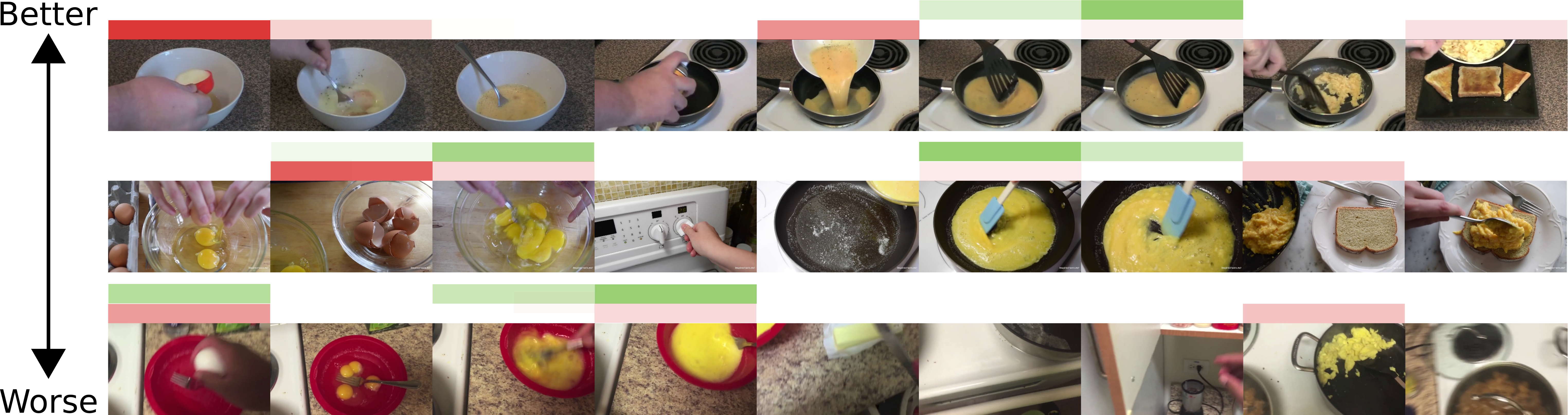}  \\
    \vspace{1em}    
    \includegraphics[width=\linewidth]{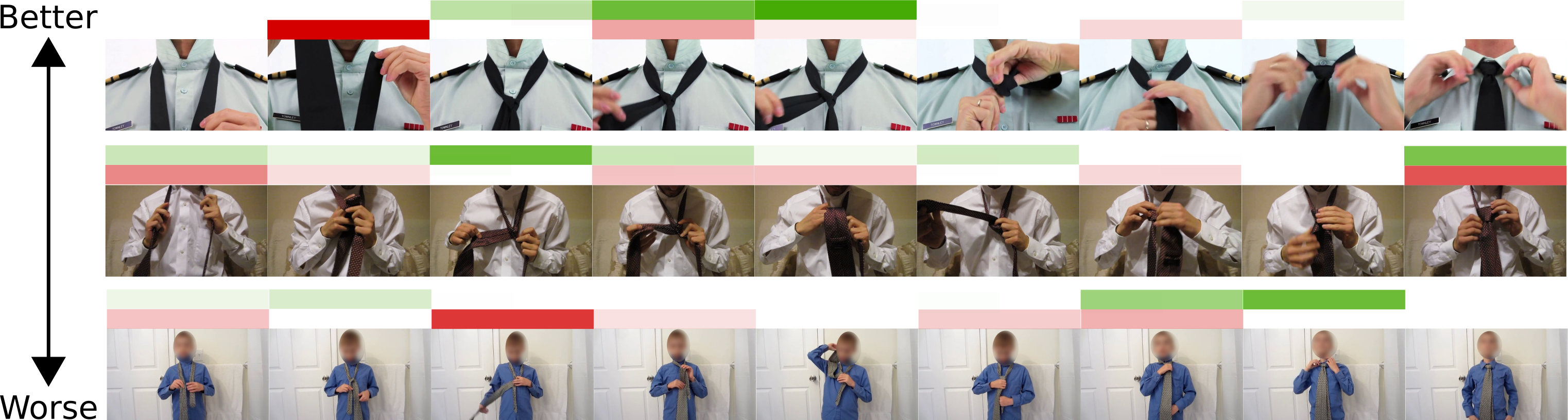}
    \caption{Attention values of the high-skill (\fgreen) and low-skill (\fred) modules with the corresponding video segments for examples from `Scramble Eggs' and `Tie Tie'. The intensity of the color indicates the attention value. We show the predicted ranking from both branches.}
    \vspace{-0.2em}
    \label{fig:qual_results}
\end{figure*}

\medskip
\noindent\textbf{Branch Contribution.}
Having trained our model with the overall loss, we now assess skill ranking using single or multiple branch scores.
From Fig.~\ref{fig:branches} we see we are able to learn high and low skill branches which are both more informative than uniform.
This is particularly true for tasks such as Chopstick Using and Scramble Eggs which see little improvement with attention until the disparity loss is introduced (Fig.~\ref{fig:ablation}). Within tasks, the performance of high and low skill branches can vary. 
We can see this for Tie Tie, with the low-skill branch performing best (+4.3\%). Here, the presence of hesitation in lower-ranked videos proves effective for skill ranking. 

The fusion of high and low skill branches further improves the result (EPIC-Skills +2.9\% and \newdataset~+3.2\%). In many tasks the branches offer complementary information, as each branch can attend to separate video segments, specific to either high or low skill (see Sec~\ref{sec:qual}).

\medskip
\noindent\textbf{Number of Filters.} In Fig.~\ref{fig:filters} we test the effect of $K$, the number of filters per attention module (Sec.~\ref{sec:att}). The previous sections report results using $K{=}3$. This shows a small improvement over one filter in the majority of tasks. 
However, with $K{>}3$ the accuracy does not increase further, as additional less-informative segments are included. 

We also compared two rank-aware attention modules, with 3 filters each, to a single standard (i.e. rank-agnostic) module containing 6 attention filters.
Results demonstrate a clear advantage of our rank-aware modules. For \newdataset, 81.2\% accuracy  drops to 75.0\% without our novel loss. 

\medskip
\noindent\textbf{Filter Correlation.}
To ensure our high and low skill filters are attending to different video segments we plot the correlation of pairs of filters between high and low attention modules, averaged over all videos for \newdataset. From Fig.~\ref{fig:correlation} we can see most filter pairs have low correlation, demonstrating these are attending to different segments. There are some cases where filters have a higher correlation (Braid Hair at $\rho=0.8$) as it can be helpful for at least one of the high and low skill filters to attend to the same segments when relevant at all levels of skill.

\subsection{Qualitative Results}
\vspace{-0.3em}
\label{sec:qual}
In Fig.~\ref{fig:qual_results} we show attention weights with corresponding frames for the Scramble Eggs and Tie Tie tasks. 
Firstly, the figure shows we are able to filter out irrelevant segments using attention, for instance turning on the stove-top and opening the cupboard in `Scramble Eggs'. Secondly, we can see our rank-aware attention allows the modules to focus on different aspects of the video. In the Scramble Eggs task the high-skill module consistently focuses on whisking the eggs and stirring the mixture in the pan,
while the low-skill module attends to adding milk/cream to the eggs and pouring. 
For `Tie Tie' the high skill module gives a strong weighting to segments displaying a tight inner knot and straightening the tie before folding across, while the low-skill module focuses mainly on hesitation and repetition. 
We also observe cases where the filters attend to segments seemingly irrelevant to skill;
in Scramble Eggs the low-skill module attends to segments containing bread. Video results are included in the supplementary material.

\section{Conclusion}
\vspace{-0.4em}
\label{sec:conclusion}
In this paper we have presented a new model for rank-aware attention, trained using a novel loss function. 
Our rank-aware loss enables us to learn the most informative segments to attend to in relation to the skill shown in the video. 
We also use the disparity loss to directly optimize the attention to pick more informative segments than the uniform distribution, solving the instability in optimizing the standard softmax attention in ranking.
We have tested this method on two datasets, one of which we introduce in this paper, and show our method achieves state-of-the-art results for skill determination, with an average performance of over 80\% in both datasets.
Future work involves exploring applications of the attention segments to improve people's skill in a task, as well as 
transfer learning to unseen tasks.

\noindent\textbf{Acknowledgements:} Access to the \newdataset~dataset and annotations available from authors' webpages. Supported by an EPSRC DTP and EPSRC GLANCE (EP/N013964/1).

{\small
\bibliographystyle{ieee_fullname}
\bibliography{egbib}
}

\end{document}